\documentclass[dvipdfmx]{interact}
\usepackage{amsmath}
\usepackage{amsfonts}
\usepackage{diagbox}
\usepackage{comment}
\usepackage{epstopdf}
\usepackage[caption=false]{subfig}

\usepackage[numbers,sort&compress]{natbib}
\bibpunct[, ]{[}{]}{,}{n}{,}{,}
\renewcommand\bibfont{\fontsize{10}{12}\selectfont}
\makeatletter
\def\NAT@def@citea{\def\@citea{\NAT@separator}}
\makeatother

\theoremstyle{plain}

\theoremstyle{definition}

\theoremstyle{remark}

\begin{document}

\articletype{SURVEY PAPER}

\title{
Explainable Autonomous Robots: A Survey and Perspective}

\author{
\name{Tatsuya Sakai\textsuperscript{a}\thanks{CORRESPONDING AUTHOR: Takayuki Nagai. Email: nagai@sys.es.osaka-u.ac.jp} and Takayuki Nagai\textsuperscript{a,b}}
\affil{\textsuperscript{a}Graduate School of Engineering Science, Osaka University, Osaka, Japan; \textsuperscript{b}Artificial Intelligence Exploration Research Center, The University of Electro-Communications, Tokyo, Japan}
}

\maketitle

\begin{abstract}
Advanced communication protocols are critical to enable the coexistence of autonomous robots with humans. Thus, the development of explanatory capabilities is an urgent first step toward autonomous robots. This survey provides an overview of the various types of ``explainability'' discussed in machine learning research. Then, we discuss the definition of ``explainability'' in the context of autonomous robots (i.e., explainable autonomous robots) by exploring the question ``what is an explanation?'' We further conduct a research survey based on this definition and present some relevant topics for future research.
\end{abstract}

\begin{keywords}
Autonomous agents; Autonomous robots; Explainability; Interpretability
\end{keywords}

\section{Introduction}



Artificial intelligence (AI) technologies have demonstrated remarkable progress and they are employed in a wide variety of applications in various fields including automatic translation, image recognition, and medical diagnosis \cite{attention,efficient,medical}. 
It is commonly claimed that AI will replace most manual labor in the future; however, is this really the case? 
AI technologies do have higher image recognition accuracy compared to humans in some limited contexts, and have consistently outperformed humans in classical games such as Go and chess.
Nonetheless, we believe that even advanced future developments based on current technology will not lead to robots replacing humans. 

AI systems' fundamental lack of ability to communicate naturally and effectively with humans is among the most significant reasons that they cannot replace human labor. Here, one may believe that such communication could be achieved via the development of natural language processing (NLP) technology \cite{BERT}; however, NLP technologies are systems for estimating the content of human statements and their meanings; they do not constitute communication. That is, humans do not feel that robots using such systems truly understand and respond to them appropriately.
Therefore, if effective communication is not achieved, robots will continue to function only as tools to assist humans. Advancements improving the accuracy or effectiveness of various specific tasks do not indicate that robots are equivalent to human beings.

Under this scenario, how can we enable robots to communicate with humans? 
One possible answer is not only to allow robots to superficially understand human instructions but also develop techniques to allow humans and robots to understand one another's internal states. 
In other words, the robot must possess the following two mechanisms.
\begin{itemize}
\item{A mechanism to estimate what users are thinking.}
\item{A mechanism that allows robots to present their own ideas in an easy-to-understand format for users.}
\end{itemize}
These mechanisms have major significance for the development of autonomous robots and their integration into human society. 
If intelligent systems can operate autonomously as instructed to assist human life, then humans should be able to understand how to use and operate them effectively. However, an autonomous robot designed to become a life partner to a human needs to convey its actions and goals in a format that is easy for the other party to understand; therefore, it must be capable of estimating the internal state of the other party and modifying its styles of actions, behaviors, and information presentation accordingly. 
Such mechanisms to enable autonomous intelligent systems to consider human are among the potential approaches by which robots may truly gain the trust of humans. 

\begin{figure}[t]
\centering
    \includegraphics[width=0.5\linewidth]{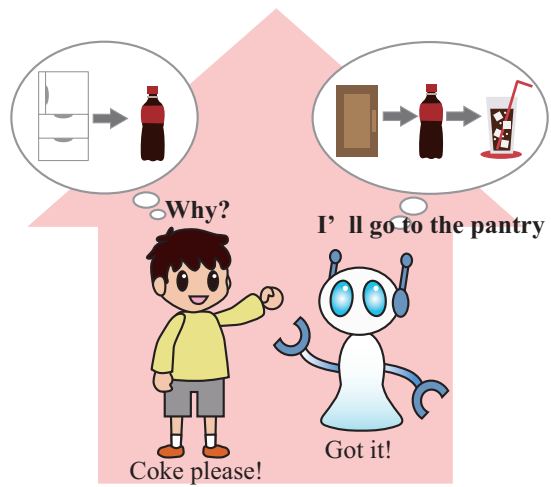}
    \caption{\small A scenario where an explainable autonomous agent is required.} 
    \label{fig:overall} 
\end{figure}

In this paper, an autonomous robot employing such a mechanism is called an explainable autonomous robot (XAR). 
We consider a specific scenario where an XAR is required, as illustrated in Fig. \ref{fig:overall}. 
Assume that a human user informs a robot to bring him a cold carbonated drink; then, as a response, the robot heads towards the pantry instead of the refrigerator. 
Potential questions that the user might ask in such a case are summarized in Table \ref{table:potential_questions}. 
%
\begin{table}[t]
\begin{center}
\caption{\small Some potential questions that the user might ask and potential answers.}
\begin{tabular}{c|c}
\hline 
Question  & Details and potential answer\\\hline 
\begin{tabular}{p{5cm}}
{\small \bf What are you doing now?}    
\end{tabular}
&
\begin{tabular}{p{7.5cm}}
{\small The user believes that the robot is moving in the direction of the refrigerator. Therefore, the question ``what are you doing now'' arises. In response to this question, the robot can say, ``I am going to get a carbonated drink from the pantry.''}
\end{tabular}\\ \hline
\begin{tabular}{p{5cm}}
{\small \bf How will this be achieved?}   
\end{tabular}
&
\begin{tabular}{p{7.5cm}}
{\small The carbonated drinks in the pantry are not refrigerated, and therefore, the user wants to ask the robot how it will provide a cold carbonated drink. As a response to this question, the robot may explain, ``I will pour the carbonated drink into a cup, and add ice.''}
\end{tabular}\\ \hline
\begin{tabular}{p{5cm}}
{\small \bf Why are you taking this particular action?}    
\end{tabular}
&
\begin{tabular}{p{7.5cm}}
{\small When the user considers that there is a cold carbonated drink in the refrigerator, they may obviously wonder why the robot is moving to the pantry to get the drink. As a response to this question, the robot can explain, ``Your younger brother drank it this morning'' (how come?) or ``So that you can have a cold carbonated drink when heading out later on'' (what for?).}
\end{tabular}\\ \hline
\end{tabular}
\label{table:potential_questions}
\end{center}
\end{table}
Failure to respond to such user questions can lead to distrust, and is a major psychological barrier in the continued use of such robot. Therefore, the ability of a robot to explain the reasoning behind its' own actions and decisions is an essential requirement for an autonomous robot. However, even if such a robot could explain all of their actions appropriately, the details would greatly surpass the amount of information that a user could accept, and this would not increase the reliability of the robot. The granularity of the necessary explanations depends not only on the information held by the user, but also on the trust relationship established between the robot and the human user. The robot must estimate how the user predicts its' behavior and present useful explanatory information to each user to achieve true explainability. In other words, the robot must estimate a model of the user and compare it with its' own behavior determination model to extract useful information for the communication of its own model and to present it to the user in a comprehensible format. 

In this paper, we conduct a survey on XARs as a first step toward developing autonomous robots that can interact with humans directly in an effective manner. A clear definition of the explainability of an autonomous robot has yet to be established, and therefore, problems associated with the explainability of autonomous robots are first defined in the following sections. Then, some significant themes are presented along with current research, followed by a discussion on issues requiring further research in future.

\section{Explainable AI (XAI) and XAR}
\label{sec:relatedarea}
In this section, we provide an overview of the existing research areas of explainable AI (XAI), explainable AI planning  (XAIP), and explainable reinforcement learning (XRL) to clarify the concept of explainability addressed in this paper. XAR as investigated in this paper is not an entirely different concept from XAI, XAIP, and XRL, and they may overlap. 
However, there are essential differences in their problem settings and goals, and therefore, it is important to provide an overview for these points.

\subsection{XAI}
A number of research projects have invested significant effort to establish a basis for the judgment of machine learning models in XAI. 
For example, joint pain was presented as the basis for diagnosing influenza, and striped patterns as the basis for identifying a zebra \cite{lime,TCAV}. 
This research area focuses on systems finding interpretations for very complex pattern recognition models such as deep neural networks. 
That is, instead of a model in which the color black is output for an input RGB value of (0,0,0), these methods envision models in which the color black is output when input is encountered that is interpreted by the system as black.
Such a model includes a factor called a ``black judgment basis'' that is not human-readable, and this factor must be presented in a human-readable format.
Such systems may play a role in close human contact work only if humans accept the presented bases of their judgments. 
Further, XAI research raises questions as to how humans perceive the reliability of such models, and expands the scope of applications of machine learning technology through the presentation of the bases of judgments.

However, XAI is not a tool to achieve human communication with autonomous agents. These techniques for presenting the basis of the judgment of a system are indeed useful; however, they are useful only for debugging the model. They do not directly contribute to research on agent decision-making in close contact with humans. 
As an example, consider the medical diagnosis of a joint pain condition mentioned earlier as an example. We are convinced when we receive an appropriate explanation, for example, ``the symptoms of joint pain contribute to the diagnosis of influenza,'' because this is consistent with our own judgment. 
If we were informed hypothetically that someone was diagnosed with influenza because of a knee scrape, rational adults would not be convinced. This is true even if the factor mentioned is an important basis for judgment. 

Thus, information presented by XAI is intended for use by debugging experts; it may not be meaningful for nonexperts. 
The explanations people would naturally seek from an autonomous robot as a partner rather than purely as a tool to assist humans would express the process by which the robot arrived at a given decision or judgment through the basis of judgment. 
In order for an autonomous robot able to perform tasks beyond a user's ability to truly communicate with the human user, it is necessary to present these process. 
The types of explainability of XAI and the XAR considered in this paper differ significantly, and therefore, we exclude research on XAI. 
For more information on XAI research, the author recommends reviewing previously published literature \cite{XAI2017,XAIsurvey2018,XAIsurvey2019,XAIsurvey2019_2,XAIbook}.

\subsection{Explainable AI Planning (XAIP)}
The major problem in XAI is the visualization of nonlinear discriminant functions in multidimensional space; the transparency regarding system decision-making and planning is not included explicitly. Further, XAIP is a research area focusing on transparency for system decision-making and planning. The explainability of autonomous agents targeted in this study is deeply related to this because of the importance of explanations related to decision-making and planning.

XAIP originated from human-aware planning (HAP) \cite{hap2014,Chakraborti2018HumanAwarePR}, which focuses on behaviors that an autonomous system such as a robot should plan when performing tasks in collaboration with humans. Planning considering humans is necessary when a person enters a control loop, as opposed to the robot performing the task alone. The basis of XAIP is the idea that humans create predictable plans for themselves by allowing for plans that humans would naturally consider rather than having to provide an explanation, removing the need for explanations \cite{Chakraborti2020TheEL}.

However, the bodies and abilities of humans and robots differ significantly, and therefore, humans and robots naturally use different global models to construct plans. Thus, robots cannot create an action plan completely similar to that of human, and in these cases, a robot may need to explain its' plan to a human. Further, XAIP is expanding its horizons to explain action plans that humans cannot imagine. In many cases, the balance between explaining and constructing a predictable plan must be considered\cite{ijcai2019-185}; this balance depends on the application. For example, communication channels and time available to a team of humans and robots in a disaster response scenario are generally limited, and there may be no time available to provide explanations. In the case of a domestic robot as mentioned above, not only are there generally expected to be sufficient communication resources, but the tasks of the robot are often different than those of a human, and thus, explanations are more important. 

XAIP is being actively studied, with workshops being held from 2018 \cite{XAIP2018WS,XAIP2019WS}, but many are considering the explanation of symbolic planning technology. In the following survey, we select some articles that present research conducted on autonomous learning agents within the field of XAIP. 
In this survey, we define XAR by referring to XAIP definition \cite{Chakraborti2020TheEL}.

\subsection{Explainable Reinforcement Learning (XRL)}
A research area referred to as XRL has introduced the idea of XAI to reinforcement learning agents \cite{puiutta2020explainable}. 
Deep reinforcement learning, which has seen remarkable development in recent years, involves high-dimensional inputs as well as a state space and policy expressed by a complex neural network; therefore, it is important to visualize the learned policy in a manner that is understandable to humans \cite{XRL2019huber,Coppens,koul2019learning}. 
That is, policy transparency is understood to be a problem setting close to XAI. 
Several proposals have been made, such as a method for replacing the policy with a program that can be read by humans \cite{pirlICML18}, and dividing a Q value into each reward source and presenting its breakdown \cite{Juozapaitis}. 
Further, causal explanations based on causal reasoning \cite{Madumal2020ExplainableRL} and the generation of contrastive explanations using counterfactual thinking \cite{madumal2020distal} have been proposed. 

The explanation of reinforcement learning agents is closely related to the explainability of the autonomous robots considered in this paper. However, the problem of policy transparency ultimately reveals what action the system learns to perform in what state, and the agent can only explain that an action should maximize the cumulative reward. 
Model-based reinforcement learning explicitly using environmental models as opposed to model-free reinforcement learning is important in terms of considering explanations while collaborating with humans using models of their behavior or wishes. 
In this paper, we survey explainability in model-based reinforcement learning. 
Considering behavioral decisions of agents based on reinforcement learning models, this is truly a planning problem, and it can be said that considering explainability overlaps with the problem of XAIP.

\subsection{XAR}
The essence of the problem considered differs between XAR and XAI. The basis of XAI is a tool-based explainability considering approaches by which tools such as AI can be of use to human work; this can also be referred to as data-driven explainability. In XAR, the problem is explaining to human the actions of an autonomous robot operating independently in direct close contact with humans. This can be referred to as goal-driven explainability, or simply as communication. 

Although XAR overlaps with research areas such as XAIP and XRL, there is insufficient organization across these areas. Additionally, new factors that have not been investigated thus far need to be considered. 
A systematic survey was conducted under the title of robot explainability \cite{XAARsurvey2019}. 
They investigated the types of papers available on a large-scale search based on keywords; however, we do not delve into considerable technical details.

\section{What is the explainability of autonomous robots?}
The explainability of autonomous robots is not a well-organized field, and exists in relation to several areas. 
This section first reviews related research on the question of the nature of such explanations. 
We then define XAR based on that knowledge.

\subsection{What is an explanation?}
There has been considerable debate regarding what exactly constitutes an explanation in various contexts. However, there is no complete answer or mathematical formulation possible for this question. Here, we review existing discussions on human explanations such as philosophy and cognitive science that are material to defining the explainability of autonomous agents in this study.

First, we need to identify whether the problem lies with explanatory process, or only with the results? Lombrozo said that explanations include both processes and results \cite{lombrozo2006structure}. An explanation is abductive reasoning that bridges the gap between the individual explaining, which in this case may be a robot, and the person to whom they are explaining; explanations can be interpreted as an action of a cognitive explanation process, or as a computational process in robots. An explanation is a dynamic process between the explaining individual and the explainee,
and therefore, it serves as an explanation not only because it crosses this gap, but because of how it does so.
The explaining individual infers the internal state of the explainee and provides an explanation that seems appropriate; the explainee then updates their own internal state. 
Further explanations are provided while inferring this state. 
In some cases, the internal state of the explaining individual is also updated. 
Thus, it can be said that this is an type of genuine communication instead of a mechanical process of simple information transmission. 
Further, it can be assumed that that perceiving the attitude of the explaining individual attempting to cross this gap of meaning will increase the sense of trust between the two. 
This attempted attitude may result in psychological agreement or alignment, even if the logical gap between the two is not fully clarified to each. 
This suggests that there are two aspects, including the content to be explained, and the approach to explanation.

If the explanation fills the gap between the explaining individuals and the explainee, it can be considered as a process wherein the explaining individual answers the questions of the individual receiving the explanation. In reference \cite{MILLER20191}, questions related to explanation are classified into three categories, as listed below.

\begin{itemize}
\item[(1)] {\bf What questions (e.g., ``What are you going to do?'')}

This is an important question, in that it is the starting point from which the two remaining questions arise. However, this question does not need to be explicit if it is clear from observable information such as movement. This question is often unnecessary between humans because of the shared body type, background knowledge, and context. In the case of XAI, the identification problem itself is shared with the user, and therefore, this question can be disregarded. However, ``what'' questions are often important for autonomous robots because of the lack of shared body type, background knowledge, or context between robots and human. Various studies have been conducted on this subject; for example, one research study visualized the behavior of a robot attempting to perform using a projection or augmented reality (AR) \cite{LiuICRA2018,HafiAR2020}.

\item[(2)] {\bf How questions (e.g., ``How will you do it?'')}

The question of how event is expected to happen can be answered using causal reasoning and causal chains \cite{Hilton2010,Hilton2005TheCO,Lagnado2008JudgmentsOC}. The ability to perform causal reasoning is important in general in autonomous robots; however, research on causal reasoning is not covered in this survey paper
because it is a very deep research area in itself.
Further, the ``how'' questions may be replaced by why questions. 

\item[(3)] {\bf Why questions (e.g., ``Why are you going to do it?'')}


From a logical point of view, ``why'' questions are probably the most difficult and the most important types of explanations. Therefore, it is very important for autonomous robots to be able to answer these questions appropriately according to the human user's intention. Dennett stated that ``why'' questions can be further divided into questions of ``how come?'' and ``what for?'' \cite{dennett1989intentional} In either case, these questions need to be answered as per the intention (context) of the questioning individual; however, it is believed that answering the question ``what for?'' is particularly important.
\end{itemize}

These discussions suggest that explanations can be considered a response to ``why'' questions \cite{MILLER20191,rosenfeld2019explainability}. That is, interpretability can be defined by how well a person can understand why a decision was made \cite{MILLER20191}. However, as mentioned at the beginning, the process is important, and therefore, it is necessary to consider what answer to give and how to answer. In the existing research on explainability, almost no research has been conducted on how to answer such a query in terms of tone, approach, and interaction style, although research has been conducted on what to answer in terms of semantic content. 
Thus, in this paper, we survey the research on the semantic content of such answers. We will touch upon social aspects such as in what manner or style to answer in the research issues section. 

To this end, the question of why people seek explanations will have significant implications if we are to consider explainability for humans. Malle cited the two following reasons why people seek explanations in the context of everyday life \cite{malle2006mind}.

\begin{enumerate}
\item {\bf Discovery of meaning:} 
Contradictions and inconsistencies between elements in the knowledge structure can be corrected through explanations.
\item {\bf Maintenance of social interaction:}
A shared meaning for things is created, and beliefs and impressions of others and emotions are changed; further, the behaviors of others are sometimes affected.
\end{enumerate}
Point (1) is important because the explanations received are utilized not only on the spot but also subsequently. It is easy to accept explanations and change one's beliefs and knowledge when it is abundantly clear that the explaining individual is providing information based on their observation to a non-explaining individual that did not make the same observation. However, it is not easy to change beliefs and knowledge for which one has a strong conviction. At the very least, the explaining individual must be fully trusted. In the case of a robot, people have common beliefs such as ``it will not lie'' and ``its knowledge is always correct''; therefore, the explanations provided by robots can function as in point (1) \cite{robottraust2012}. However, considering the interaction between explanations, the robot itself needs to be able to modify its own knowledge structure while communicating explanations; this remains a difficult problem. The reliability of the information, the other party, and one's own knowledge all need to be considered.

Point (2) suggests the importance of considering the social aspect of an explanation. This problem of maintaining relationships is associated with the interactive process of the explanations mentioned at the beginning and the overall approach to explanation. People's beliefs towards robots as described in point (1), including ``robots do not lie'' and ``robot knowledge is correct,'' change based on interactions. Therefore, we need to consider approaches to explaining such that this belief can be maintained. However, the current status of research on the social aspects in explanations by AI and robots remains limited or lacking. 

References \cite{lombrozo2006structure} and \cite{Wilkenfeld2015} state that explanations not only convey certain types of knowledge, but they also have functions such as persuasion, learning, and in some cases, distraction. In a social context, the purpose is often not shared between the explaining individual and the individual to whom they are explaining. This is natural when considering that the explanation is ultimately the communication itself as opposed to only the semantic content thereof. Instead, it can be said that this subject is related to the field of human–robot interaction (HRI), and thus far, no research related to XAR has pursued these points.

\subsection{Preparation of defining explainability}
%
\subsubsection{Decision making}
A decision-making space $\Pi$ with Markov properties is expressed using a transition function
\begin{equation}
\delta_{\Pi}:A \times S \rightarrow S \times \mathbb{R},
\end{equation}
where $A$ denotes a set of selectable actions, $S$ represents the set of transition-feasible states, and $\mathbb{R}$ denotes the set of all real numbers representing the cost required for transition. At this point, the decision-making problem of an autonomous agent can be formulated as an algorithm $\mathbb{A}$ generating a plan or policy $\pi$ under a certain constraint $\tau$ (index such as optimality or explainability) in the decision-making space $\Pi$ \cite{Chakraborti2020TheEL}. That is,
\begin{equation}
\mathbb{A} : \Pi \times \tau \rightarrow \pi.
\end{equation}
The plan here can be expressed as
\begin{equation}
\pi = <a_1, a_2, \cdots, a_n>,~~~ a_i \in A.
\end{equation}
Further, the policy can be expressed as
\begin{equation}
\pi : s \rightarrow a,~~~\forall a \in A,~~~ \forall s \in S.
\end{equation}
In the following, the words ``plan'' and ``policy'' are used interchangeably, 
because the plan can be derived from the policy
\footnote{We assume that the robot has a model of the environment (world model).}. 

\begin{figure}[t]
\centering
    \includegraphics[width=1\linewidth]{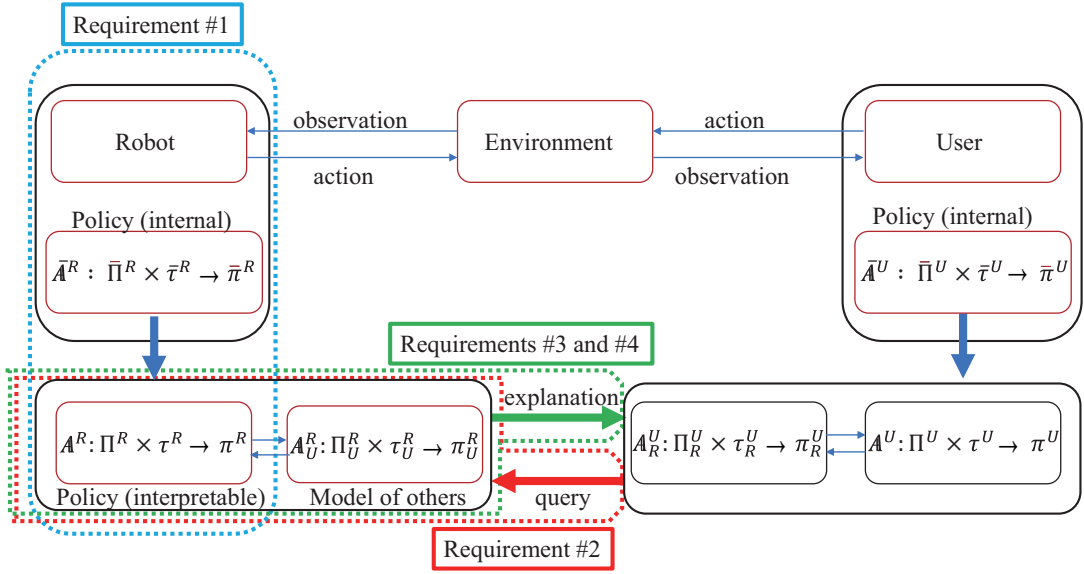}
    \caption{\small 
    Explanations provided to the user of the robot assumed in this paper.} 
    \label{fig:detailed_model} 
\end{figure}
%
\subsubsection{Overall structure of explanations}
The overall structure of explanations assumed in this paper using the definitions provided above is illustrated in Fig. \ref{fig:detailed_model}. 
First, both the robot and the human user have internal policies, and they both act accordingly. That is, we assume that the above decision-making process can be applied not only to robots, but also to humans. These internal policies are not directly observable from the outside or by the individuals themselves; therefore, they need to be converted to or approximated by an interpretable policy to interpret or explain one's internal policy to others. These interpretable policies are expressed as 
\begin{equation}
\mathbb{A}^{R} : \Pi^{R} \times \tau^{R} \rightarrow \pi^{R}~~:for~robot,~~~~\mathbb{A}^{U} : \Pi^{U} \times \tau^{U} \rightarrow \pi^{U}~~:for~user.
\end{equation}
Furthermore, this includes a model of others, which is an estimate of the interpretable policy of the other party. This implies that $\mathbb{A}_{R}^{U}$ (the human user's cognitive model of the robot's behaviors, goals, and decision-making tendencies) is an estimation of $\mathbb{A}^{R}$, and $\mathbb{A}_{U}^{R}$ (the robot's computational model of the human user's behaviors, wishes, and preferred interaction styles) is the estimation of $\mathbb{A}^{U}$.

If we assume that the estimation of each model is accurate, the explanation provide by the robot to the user is generated using information regarding the interpretable policies of the robot and the user; this in turn affects the interpretable policy of the user, and further affects the internal policy of the user. 

Here, it is assumed that the user and the robot are able to perform the same task, and that there is no difference in physical or computational abilities. In this case, the interpretable policies of the user and the robot should match, and the explanation process can then be formulated from the difference between the interpretable policy and the model of others. In cases where there are differences in tasks and abilities, $\mathbb{A}^{U}$ is considered the interpretable policy that the user expects from the robot, instead of the user's own interpretable policy. Therefore, $\mathbb{A}_{U}^{R}$ is the users' expected interpretable policy of the robot as estimated by the robot. 
In this case, the formulation is the same, and we therefore assume that the abilities of the human and robot considered to perform tasks are identical\footnote{The problem of estimating the expected interpretable policy is more difficult for the robot than estimating the interpretable policy of the user.}.

\subsubsection{When an explanation is required}
For explanations that need to answer ``why?'' questions, we formulate the problem by considering the types of scenarios that require an explanation. ``Why'' questions can be written as ``why $\pi$?'' Or ``why not $\hat{\pi}$?'' (or partially, ``why $a_i$?'' / ``why not $\hat{a_i}$?''). These ``why'' questions occur in the following scenarios.
\begin{equation}
\mathbb{A}_{R}^{U} : \Pi_{R}^{U} \times \tau_{R}^{U} \rightarrow \pi_{R}^{U}~~and~~\mathbb{A}^{U} : \Pi^{U} \times \tau^{U} \rightarrow \pi^{U} \neq \pi_{R}^{U}.
\label{eq:user}
\end{equation}
Equation (\ref{eq:user}) is a process inside the user and cannot be directly handled by the robot. Therefore, the robot requires explanations in the following scenarios.
\begin{equation}
\mathbb{A}_{U}^{R} : \Pi_{U}^{R} \times \tau_{U}^{R} \rightarrow \pi_{U}^{R}~~and~~\mathbb{A}^{R} : \Pi^{R} \times \tau^{R} \rightarrow \pi^{R} \neq \pi_{U}^{R}.
\label{eq:askwhen}
\end{equation}

$\pi_{U}^{R}$ needs to be brought closer to $\pi^{R}$ by answering ``why'' questions to eliminate the discrepancy between the user and the robot\footnote{However, this is a problem on the user side, and therefore, $\pi^{U}$ needs to be brought closer to $\pi_{R}^{U}$ in practice.}. 
Equation (\ref{eq:askwhen}) includes the degrees of freedom on both the user and robot sides, and thus, several methods can be considered to bring the two policies closer to one another. 
For example, in XAIP, the main point is to adjust $\tau^{R}$ for changing the robot's plans and bring $\pi^{R}$ closer to $\pi^{R}_{U}$. 
XRL can visualize $\pi^R$ to display information, which allows the user to understand $\pi^R$. 
In a debug scenario, the algorithm $\mathbb{A}^{R}$ should be visualized so that the designer can understand the system well. 
In XAR, the main focus is on the discrepancy of $\Pi$; eliminating this discrepancy in $\Pi$ can be achieved by using the methods involving changing one's own $\Pi^R$ and providing information (explanation) for changing the user's $\Pi^{U}$, and a method combining both these approaches. We consider the method of providing information for changing the user's $\Pi^{U}$.

\subsection{Definition of explanation in XAR}

Let the explanation of the robot to the user be denoted by $\epsilon$. Based on the above formulation, the generation of $\epsilon$ in the following equation by the robot can be defined as an explanation generation process \cite{Chakraborti2020TheEL}.
\begin{equation}
\begin{split}
\label{eq:explanation}
& Given: \mathbb{A}^{R} : \Pi^{R} \times \tau^{R} \rightarrow \pi^{R} \\
& \Pi_{U}^{R} + \epsilon \rightarrow \hat{\Pi}_{U}^{R} ~~such~that~~ \mathbb{A}_{U}^{R} : \hat{\Pi}_{U}^{R} \times \tau_{U}^{R} \rightarrow \pi^{R}.
\end{split}
\end{equation}

In reference \cite{model_of_other_PDDL}, this approach is called the model reconciliation problem (MRP). 
This problem is then investigated using the planning approach described the planning domain description language (PDDL) \cite{PDDL98}.	
	
The important aspect in this explanation generation is that $\epsilon = \pi^{R}$ does not hold. Such a direct method is one approach to XAI, however, directly presenting $\pi^{R}$ does not answer the ``why'' question, as is evident from the discussions provided thus far. Moreover, $\epsilon$ must be conveyed to people as information in practical scenarios. In the context of XAR, the following requirements must be met for generating an explanation $\epsilon$ in Eq. (\ref{eq:explanation}) and for explaining it to a user.

\begin{itemize}
\item[\bf 1)] {\bf The autonomous robot owns an interpretable decision-making space $\Pi$:}

As mentioned above, the internal decision-making spaces of others obviously cannot be accessed, and therefore, an interpretable decision-making space needs to be maintained\footnote{
  The robot can directly use the internal decision-making space as the interpretable decision-making space in some cases
depending on the implementation form of the internal decision-making space.}. The important point here is identifying whether the interpretable decision-making space is comprehensible to the human user, which implies that its' meaning is interpretable. Each state transition corresponds to the smallest unit of decision making and plays the role of an atom in symbolic reasoning.

\item[\bf 2)] {\bf The $\mathbb{A}_R^U$, $\Pi_R^U$, and $\tau_R^U$  by the user (model of others) are estimated:}

The explanation is an adjustment of the differences between a robot or human agent's own interpretable policy and the interpretable policy of the user to be communicated with, and therefore, the interpretable policy of the user (model of others) needs to be estimated\footnote{The optimal content of the estimation of the model of the other depends on an assumption; if $\mathbb{A}^U$ and $\tau^{U}$ are assumed to be shared, then the target to be estimated is $\Pi_R^U$.}.

\item[\bf 3)] {\bf Information necessary for the user to estimate $\pi_R$ needs to be estimated:}

Explanation $\epsilon$ in Eq. (\ref{eq:explanation}) needs to be estimated from one's own interpretable policy and the estimated model of others.

\item[\bf 4)] {\bf Means of presenting explanations to users:}

The explanation $\epsilon$ generated by requirement 3 needs to be encoded into languages and/or images and conveyed to a person.

\end{itemize}

We believe that satisfying the above requirements can define the explainability required by an autonomous robot. 
Requirements 1 and 2 are necessary for generating explanations, and requirements 3 and 4 are phases in which the models established are used to construct explanations\footnote{Different classifications can be considered for requirements 3 and 4. For example, in references \cite{neerincx,XAARsurvey2019}, three phases of explanation were considered in the context of communication, including (i) generation of an explanation inside an agent, (ii) transmission of explanation in a form that is easily accepted by the user, and (iii) acceptance of explanation by the user. Reference \cite{Chakraborti2020TheEL} classified methods of explanation as (a) decision-making algorithm-based explanations with the objective of debugging, (b) model-based explanations designed for compensating for differences in cognitive or computational agent models and associated information-processing capabilities, and (c) plan-based explanations for conveying a series of action plans.}. The details of these requirements are provided in the following subsections. 


\subsection{Owning an interpretable decision-making space $\Pi$: Requirement 1}
The user's interpretable decision-making space $\Pi$ is expressed as
\begin{equation}
\begin{split}
\label{spacedef}
&Given:\mathbb{A}^U,\ \tau^U  \\
&\Pi\ such\ that\ \mathbb{A}^U : \Pi \times \tau^U \not\rightarrow \phi,
\end{split}
\end{equation}
where $\mathbb{A}^U$ denotes the user's planning algorithm, and $\tau^U$ denotes a type of constraint. When a policy is derived using these and $\Pi$, and when some policy $\pi$ is determined without encountering an error, the decision-making space $\Pi$ is defined as interpretable to the user. 
This definition implies that $\Pi$ is composed of states and actions that are natural to humans, and the transition between states is consistent with the conventions that govern the real world (e.g., laws of physics)\footnote{Such an interpretable decision-making space can be acquired by learning, including physical identity and interaction. 
The existence of this interpretable decision-making space is related to the encoding of meaning in languages, and it can be explained by the idea of symbol emergence in robotics \cite{SER2015,SER2019}, which is a mechanism by which a symbol system is socially constructed and shared.}. 

The easiest approach to build a user-interpretable decision-making space is for humans to write the rules. 
However, as is well known, this approach has limitations in creating such systems and eventually faces the problem that humans cannot write a complete rules set determining a single correct action in every conceivable situation. 
Further, another approach has also been developed involving constructing a causal graph from certain information to the target state \cite{madumal2020distal}. 
This research is actively being pursued in a field called ``causal discovery''\cite{causaldiscovery2019}; however, it is considered difficult to model the vast number of factors in the world sequentially and achieve a framework that can comprehensively learn these causal relationships. For the construction of causal graphs, the factors of interest are specified by humans, and it is difficult to say that they constitute autonomous learning by the system.

An alternative to the above-mentioned methods is to build a decision-making space that can be interpreted by humans; this can be achieved by having the robot itself acquire a world model \cite{worldmodel}. 
The world model framework involves robots modeling the relationship between their own behavior and changes in the outside world through interactions with their environment. 
Information on changes in the outside world is obtained by observations by the robot itself, and therefore, the decision-making space of a robot can be understood by humans provided the system includes a mechanism for converting the observed information of the robot into a form that can be interpreted by humans. 
The world model framework is similar to the conventional symbolic reasoning framework, but does differ signficantly. 
It has the potential to build a decision-making space that humans can understand without direct human intervention through deep learning and advanced state representation learning methods \cite{SRLforControl2018}\footnote{Inference without action can also be modeled conceptually with a similar approach. 
For example, when considering the transition from the ``state of being infected by the influenza virus'' to a ``joint pain state,'' there is a transition from the ``state of being infected by the influenza virus'' to ``prostaglandin is generated'' because of the immune reaction. This substance then increases the body's responsiveness to pain and reaches the ``joint pain state.''}. 

One challenge to learning in such a real-world model is that actions need to be pre-defined. 
When modeling state changes accompanying the actual action of an agent, the world model can be constructed at the motor command level of the robot. 
In an environment where actions cannot be easily defined, a framework is required in which even selectable actions are learned autonomously\footnote{For example, methods such as action segmentation \cite{nakamura2017seg,nagano2019} can be used.} and meaning is assigned to actions based on differences between states before and after an action.

\subsection{Estimation of the model of others: Requirement 2}
In the MRP, solving Eq. (\ref{eq:explanation}) and generating an explanation $\epsilon$ requires that $\mathbb{A}^R_U \approx \mathbb{A}^U$, $\Pi^R_U \approx \Pi^U$, and $\tau^R_U \approx \tau^U$ be satisfied. Here, if $\mathbb{A}^R_U \approx \mathbb{A}^U$ and $\tau^R_U \approx \tau^U$, then the problem is estimating $\Pi^R_U$.
\begin{equation}
\begin{split}
\label{eq:requirement2}
& Given: \mathbb{A}^{R} : \Pi^{R} \times \tau^{R} \rightarrow \pi^{R},~ \mathbb{A}^R_U \approx \mathbb{A}^U,~ \tau^R_U \approx \tau^U \\
& \Pi_{U}^{R}  \rightarrow \hat{\Pi}_{U}^{R} -q  ~~such~that~~ \mathbb{A}_{U}^{R} : \hat{\Pi}_{U}^{R} \times \tau_{U}^{R} \rightarrow \pi^{R}.
\end{split}
\end{equation}
Here, $q$ represents a query from the user. 
The estimation of $\mathbb{A}^U$ or $\tau^U$ can be considered a more general problem; however, a comprehensive estimation is believed to be a very difficult problem. Even if the decision-making space is identical to that of the world model held by the human, the planning algorithm does not indicate the reward function in decision-making. The size of the decision-making space that humans can consider when planning, the observable information, and other operators from the outside world are entangled and interdependently conditioned in a complex manner and thus difficult to determine via a computational mechanism.

Estimating the decision-making space and planning algorithms of humans plays a very important role in explanation generation. For example, it would seem obvious that a key is required to open a front door to a home, but this may not be obvious to a toddler whose parents always open the door for them. When instructing such an agent lacking contextual information to ``open the door,'' one could say, ``place this key in the keyhole, turn it to the left, and then pull the door.'' Human beings consider such factors daily, and this task captures the essential meaning of ``communication.''
\footnote{This example was related to a behavioral sequence; however, this applies to knowledge reasoning as well. For example, if someone did not understand the connection between joint pain and influenza, it might be better to explain that ``a fever occurred to get rid of the influenza virus, and the substance secreted in the body at this time causes joint pain.'' However, to do so, not only is the construction of a decision-making space using the information given by humans necessary, but the system must also autonomously and widely learn more detailed information and seemingly unrelated information. From this point of view, the autonomous acquisition of the abovementioned decision-making space is indispensable in explanation generation.}


\subsection{Estimation of the information needed for a user to estimate $\pi^R$: Requirement 3}
This requirement is the problem of identifying a method to find $\epsilon$ in Eq. (\ref{eq:explanation}). Generating explanations that promote human comprehension of robot policies in the decision-making space requires explanations that are easily accepted as (i) having some human characteristics and (ii) appropriate for specific individuals. These two points are discussed in the following sections.

\subsubsection{Good explanation from the perspective of cognitive science}
We consider explanations that are easily accepted by humans from the perspective of cognitive science. According to cognitive science, two approaches are understood to serve as indicators to determine the quality of an explanation, including likeliness and loveliness \cite{explanatory,lipton}. Likeliness is a probabilistically defined indicator of explanation quality. In the likeliness framework, a good explanation is defined one that maximizes the posterior probability $P(X|E_i)$ of an event $X$ that we wish to explain when a certain explanation $E_i$ is presented. Loveliness, in contrast is an indicator defined from an axiological perspective. 
In the framework of loveliness, studies have focused on simplicity and latent scope as indicators that determine the goodness of an explanation.
Simplicity is an index expressing the number of assumed causes, where one with a smaller number is preferred by humans \cite{lombrozo}. Latent scope was predicted to be introduced by a cause, and this was ultimately an unobserved event; a higher posterior probability $P(X|E_i)$ was estimated with a smaller number for the latent scope \cite{harrypotter,Johnson}.

According to the likeliness and loveliness indicators, the minimum information necessary for comprehension should be provided when an autonomous agent explains the reason for making a decision. In other words, factors important for bringing cognitive or psychological policies of humans and computational policies of machine agents together to a certain extent should be provided. At this point, the tolerance in the route assumed by an agent and that assumed by the human may not always be constant. Further, there are cases where we are satisfied if we understand that an agent will reach its target state, and there are other cases where we want to understand the specific route that the agent took in its decision-making space accurately. The balance between the simplicity and accuracy of an explanation varies greatly depending on the quality of explanations required.

\subsubsection{Good explanations for individuals}
The quality of an explanation for a specific individual can be understood from the perspective of cognitive science. However, simply presenting ``a generally good explanation'' to all users is insufficient because
\begin{itemize}
\item{the decision-making space of each individual is different,}
\item{differences in planning algorithms change the state transition sequence estimated by humans in the decision-making space before receiving an explanation,}
\item{the method and accuracy of reflected information obtained from the explanation in one's own model differ significantly.}
\end{itemize}
When an autonomous agent presents an explanation to a specific individual, it should first estimate the relevant decision-making space and planning algorithm, after which it should find important elements for policy transmission based cognitive science in the intersection of the computational agent's and human's decision-making spaces. If the intersection cannot generate a route from the initial state to the target state, or if this becomes extremely complicated, the agent or the human needs to supplement the lack of knowledge or correct the misidentified knowledge. 

%
\subsection{Presentation to user of explanation: Requirement 4}
Simply solving Eq. (\ref{eq:explanation}) does not result in an explanation. The robot needs to encode the information, i.e., $\epsilon$, into a form that can be conveyed to humans. The problem in Eq. (\ref{eq:explanation}) can be strictly written as 
\begin{equation}
\begin{split}
\label{eq:explanation_full}
& Given: \mathbb{A}^{R} : \Pi^{R} \times \tau^{R} \rightarrow \pi^{R} \\
& \Pi_{U}^{R} + dec(enc(\epsilon)) \rightarrow \hat{\Pi}_{U}^{R} ~~such~that~~ \mathbb{A}_{U}^{R} : \hat{\Pi}_{U}^{R} \times \tau_{U}^{R} \rightarrow \pi^{R}.
\end{split}
\end{equation}
Here, $enc(\cdot)$ denotes the function that encodes $\epsilon$, and $dec(\cdot)$ denotes the function that receives the encoded information and decodes it. Methods such as verbalization \cite{Hayes,Waa,Rosenthal2017} or visualization \cite{LiuICRA2018,HafiAR2020} can be considered for $enc(\cdot)$. 
This is a more difficult issue, and it includes the  modality in which the information should be encoded, and simultaneously, there is a potential of creating new explanatory media specific to robots.

\section{Survey on key issues in XAR and associated research}
Issues for achieving an explainable robotic agent can be divided into four points listed in Table \ref{table:four_requirements} based on the discussions provided above. 
\begin{table}[t]
\begin{center}
\caption{\small Four requirements for achieving XAR.}
\begin{tabular}{c|c}
\hline 
Requirement  & Details\\\hline 
\begin{tabular}{p{4cm}}
{\small {\bf Requirement \#1:} A robot can autonomously acquire a decision-making space that is interpretable by humans (a space where each decision can be interpreted by humans)}    
\end{tabular}
&
\begin{tabular}{p{8.5cm}}
{\small This implies that each individual decision is interpretable by humans. Although introducing a world model framework is effective, it is preferable to build a framework that can autonomously learn the set of selectable actions.}
\end{tabular}\\ \hline
\begin{tabular}{p{4cm}}
{\small {\bf Requirement \#2:} Estimating a human decision-making space and planning algorithm}
\end{tabular}
&
\begin{tabular}{p{8.5cm}}
{\small Autonomous agents first estimate the human decision-making space and planning algorithm to share their own intentions and thoughts with humans. The planning algorithm does not necessarily indicate the reward function in the action decisions, which is determined by the intricate intertwining of various factors, such as the size of the decision-making space that humans consider when planning, observable information, and other operators from the outside world.}
\end{tabular}\\ \hline
\begin{tabular}{p{4cm}}
{\small {\bf Requirement \#3:}  Extracting information that is important for communicating policies}
\end{tabular}
&
\begin{tabular}{p{8.5cm}}
{\small The error between the route assumed by agents and that assumed by humans is held within a certain tolerable range; alternatively, useful information is extracted and explanatory factors are determined to reduce the estimated load on humans.}
\end{tabular}\\ \hline
\begin{tabular}{p{4cm}}
{\small {\bf Requirement \#4:}  Converting explanatory factors into the most efficiently transmitted form (verbalization or visualization of explanations)}
\end{tabular}
&
\begin{tabular}{p{8.5cm}}
{\small The extracted explanatory factors are converted into a form that is interpretable by humans, such as language or visual expressions. A mechanism needs to be developed to flexibly modify the expression of explanations after considering the method and accuracy of reflected information obtained from explanations by humans in their own models.} 
\end{tabular}\\ \hline
\end{tabular}
\label{table:four_requirements}
\end{center}
\end{table}
\begin{figure}[t]
\centering
    \includegraphics[width=0.8\linewidth]{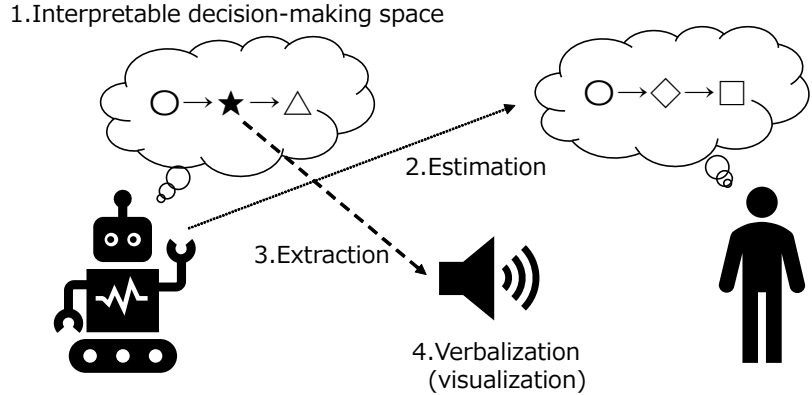}
    \caption{\small Schematic of the research issues.} 
    \label{fig:survey_overall} 
\end{figure}
Figure \ref{fig:survey_overall} shows a schematic of these research issues. 
We classified existing research on the explainability of autonomous agents based on whether they meet the above requirements (Table\ \ref{table:previous_method}). 

\begin{table}[t]
\begin{center}
\caption{\small Existing methods and research target requirements. The checkmarked content is not always the subject of the article because the classification is based on the research issues.}
\begin{tabular}{c|c c c c c}
\hline
Method $\backslash$ Requirement  & 1  &  2 &  3 & 4 & other\\\hline 
Wang {\it et al.} \cite{Wang} & \checkmark   &  &  & &\\
Verma {\it et al.} \cite{pirlICML18} &  \checkmark  &    &    & &\\
Coppens {\it et al.} \cite{Coppens} & \checkmark  &   &   & &\\
Zhang {\it et al.} \cite{Zhang} & \checkmark  &   &   & &\\
Gopalakrishnan {\it et al.} \cite{Gopalakrishnan} & \checkmark  &   &   & &\\
Sakai {\it et al.} \cite{methodXAR} & \checkmark  &   & \checkmark  & &\\
Clair {\it et al.} \cite{Clair}  &   & \checkmark &  & &\\
Gao {\it et al.} \cite{Gao}  &   & \checkmark &  & &\\
Huang {\it et al.} \cite{Huang} &    & \checkmark & \checkmark & &\\
Lage {\it et al.} \cite{Lage} &    & \checkmark & \checkmark & &\\
Khan {\it et al.} \cite{Khan}  &   &  & \checkmark &  &\\
Dodson {\it et al.} \cite{Dodson}  &   &  & \checkmark & &\\
Amir {\it et al.} \cite{Amir}  &   &  & \checkmark & &\\
Madumal {\it et al.} \cite{madumal2020distal} &    &  & \checkmark & &\\
Sequeira {\it et al.} \cite{Sequeira} &    &  & \checkmark & \checkmark &\\
Hayes {\it et al.} \cite{Hayes}  &   &  &  & \checkmark &\\
Waa {\it et al.} \cite{Waa} &    &  &  &\checkmark &\\
Ehsan {\it et al.} \cite{Ehsan}  &   &  &  & \checkmark &\\
Yeung {\it et al.} \cite{Yeung}  &   &  &  & \checkmark &\\
Huber {\it et al.} \cite{Huber}  &   &  &  & \checkmark &\\
Das {\it et al.} \cite{Das}  &   &  &  & \checkmark &\\
Elizalde {\it et al.}   \cite{Elizalde}  &   &  &  & & \checkmark \\
Dragan {\it et al.} \cite{Dragan}  &   &  &  & & \checkmark \\
Fukuchi {\it et al.} \cite{Fukuchi}  &   &  &  & & \checkmark \\
Juozapaitis {\it et al.} \cite{Juozapaitis}  &   &  &  & & \checkmark \\
\hline
\end{tabular}
\label{table:previous_method}
\end{center}
\end{table}

\subsection{Autonomous acquisition of interpretable decision-making space}
Building a human-interpretable decision-making space and making decisions in that space is the basis for estimating the human decision-making space and for generating explanations acceptable to humans. Wang {\it et al.} \cite{Wang} showed that, in partially observable Markov decision process (POMDP), where the input is an interpretable element set, the certainty of state transitions could be communicated to humans not only by presenting the information after the transition but also by supplementing the defects in the robot's own capabilities such as its sensors and the reliability of its available information. 
Verma {\it et al.} \cite{pirlICML18} constructed an interpretable policy of an agent by approximating the policy programmatically under constraints given by humans. 
In addition, Coppens {\it et al.} \cite{Coppens} assumed that input features were interpretable and numerically descriptive, and they provided single decisions that were interpretable by a decision tree using important features. 
These methods make each decision as interpretable as possible by assuming that they can decompose the inputs used for decision-making into interpretable features and that these decomposition methods are known. Thus, although an interpretable decision-making space is constructed, as yet no method has been developed for constructing a decision-making space that can be applied universally in any environment without human intervention.

In research on world models, Zhang {\it et al.} \cite{Zhang} identified and graphed landmarks based on the proximity of representational features, enabling agents autonomously to acquire abstract decision spaces in any environment. 
In addition, Gopalakrishnan {\it et al.} \cite{Gopalakrishnan} proposed a state space abstraction method reducing the number of branching points in an action strategy with the aim of increasing the predictability of a computational  agent's actions. 
However, the decision spaces generated by these methods are not always interpretable by humans, and it is necessary for humans to give meaning to the results obtained. 
We previously proposed a method to generate explanations using an autonomously acquired world model \cite{methodXAR}. 
However, it is only designed used in environments where observations are discrete and interpretable, and many issues remain that should be considered regarding XAR.

\subsection{Estimation of human decision-making space and policy}
Estimating a human decision-making space and its associated policies is an essential requirement for generating personalized explanations. 
Clair {\it et al.} \cite{Clair} assumed that humans and robots held identical sets of policies and proposed a framework for estimating plausible policies from human actions. 
Gao {\it et al.} \cite{Gao} proposed a framework for estimating a plausible policy currently envisioned by a user, considering not only the user's behavior but also an interaction history.
Huang {\it et al.} \cite{Huang} prepared multiple definitions of policy estimation methods and plausibility in inverse reinforcement learning, and showed that policies reproduced from the presentation of the same information differed based on the definition used. 
Lage {\it et al.} \cite{Lage} showed that differences in human policy restoration models changed the accuracy of a task of restoring behavioral sequences from summaries. 

These studies not only suggested methods for estimating human policy but also argued for the importance of modeling humans who receive explanations. However, as yet no models have been proposed that perform well on the task of estimating a human internal model.

\subsection{Extraction of key elements in the process}
Even if an interpretable decision-making space can be constructed, presenting a decision-making process from that space does not make it an explanation that can be easily accepted by humans. 
We can generate explanations that are acceptable to humans only by extracting the elements important for understanding the decision-making process in that space. 
Huang {\it et al.} \cite{Huang} and Lage {et al.} \cite{Lage} assumed that humans were using a framework of inverse reinforcement learning and imitation learning to restore policies, and the element with the highest restoration accuracy was extracted as the important element. 

Khan {\it et al.} \cite{Khan} assumed a cognitive model based on Markov decision processes (MDPs), and they obtained the ``best expected reward value'' and ``worst expected reward value'' at the time of action selection in a given state; further, they extracted the state where the difference between reward values is large as the important factor. 
Dodson {\it et al.} \cite{Dodson} showed the state in which an action value at the next time period became particularly high when the action was fixed to an optimum as important factor, in addition to explanation using learning data.
Amir {\it et al.} \cite{Amir} calculated the difference between the maximum and minimum action values in each state in a MDP, and elements with the high values were presented as scenes constituting good representations of agent characteristics. 
Sequeira {\it et al.} \cite{Sequeira} extracted states with a high frequency of appearance and those with a high variance of action selection frequency; these scenes were summarized as a video clip. 
Madumal {\it et al.} \cite{madumal2020distal} generated explanations by projecting a node in an action decision tree of a depth-constrained agent onto a causal graph of input features rather than by an MDP.

As shown above, many studies have been conducted on the identification of important factors among the requirements for explainability in autonomous agents. These studies mainly sought to present important information to humans by extracting states where the difference in the action value in an MDP was large. Therefore, extractions cannot be made by such methods in cases where there are essential requirements for reaching the target state despite no major differences in action value. 

In reference \cite{methodXAR}, we identified important factors by approximate calculation of the causal effect on the probability of reaching a target state without using an action value. This method was shown to be able to extract important elements for reaching a target state regardless of the action value, but problems remain regarding its computational complexity and optimal hyperparameter settings.

\subsection{Verbalization and visualization of explanations}
Verbalizing and visualizing important elements and reducing the burden of information interpretation are important to generate explanations that can be easily accepted by humans. 
Hayes {\it et al.} \cite{Hayes} verbalized an agent's policy by finding a set of states in which the agent selected an action in an MDP; they presented a linguistic explanation set expressing that state. 
Waa {\it et al.} \cite{Waa} generated explanations by inferring the state and results to be reached from actions taken in response to a question posted as ``Why was $a^+$ and not $a^-$ selected?'' and by presenting the language connected to this in advance.
Ehsan {\it et al.} \cite{Ehsan} and Das {\it et al.} \cite{Das} proposed a framework for directly generating linguistic explanations from agent state sequences using an encoder-decoder model.

Sequeira {\it et al.} \cite{Sequeira} presented important scenes as video clips and aimed to generate human-acceptable explanations. 
Huber {\it et al.} \cite{Huber} applied a saliency map, which was conventionally used to improve the interpretation of image classification tasks, and attempted to improve the interpretation of behavioral strategies by highlighting important parts of an image.

These methods make it possible to present information using interfaces such as language and images that are easy for humans to understand. However, in interpersonal explanations, it is desirable to change not only the information to be explained but also the manner in which the explanation is presented according to the user. 
Therefore, Yeung {\it et al.} \cite{Yeung} proposed a method to select the best available explanation presentation method by incorporating the process of explanation presentation and user understanding into a reinforcement learning framework. 
In order to apply the method to real-world problems, further discussion on reward design and training data acquisition methods would be beneficial.

\subsection{Other research studies}
Some of the papers reviewed as contributing to the realization of XAR do not fit into any of the four above categories. 
The four requirements that have been addressed so far are essential for the realization of XAR, but it may be possible to generate a more suitable explanation for a given situation by using the methods shown below together as necessary. 
\begin{description}
    \item [Elizalde {\it et al.} \cite{Elizalde}]\mbox{}\\
    extracted important features from an input by finding the effect of each feature on the utility function and their effect on the selection action. 
    This method applies the idea of XAI to RL, and can present the features that are emphasized in decision-making in each state.
    \item [Dragan {\it et al.} \cite{Dragan}]\mbox{}\\
    designed a route supporting the prediction of goals from the outset for humans as opposed to generating explanations for actions. 
    This method was based on the concept of XAIP, and remains an important approach to ease the human burden of understanding an explanation.
    \item [Fukuchi {\it et al.} \cite{Fukuchi}]\mbox{}\\
    set a goal of the difference 
    between the state at the current time $t$ and some future time $t+n$ when the action was decided by an optimal policy, and they presented the most plausible explanation for that goal. 
    We consider the meaning of an action based on the difference in state is more interpretable than the meaning at the motor command level, and is useful for explanation. 

    \item [Juozapaitis {\it et al.} \cite{Juozapaitis}]\mbox{}\\
    conducted reinforcement learning by dividing the Q-value of each reward source, and presented the breakdown. 
    When multiple reward sources were assumed, a brief explanation can be provided simply by showing the important reward sources. 
\end{description}

\section{Future research issues}
There are many requirements for realizing an XAR; however, there are several areas where the research is already underway and others wherein these requirements as yet have hardly been considered. Further, in some areas, the current state-of-the-art methods encounter a considerable number of issues regarding the realization of an agent that can autonomously interact with humans in the real world. In this section, we discuss the primary future research issues as the conclusion of this study. 

\subsection{Estimation of decision-making space and planning algorithm of others}
Appropriately estimating the internal state of a conversation partner is an essential requirement for realizing an XAR; however, the current state is such that this framework has not yet been realized. 
The definition of ``understanding the other'' is not obvious; further, it is believed that the information to be modeled depends highly on the assumed domain. However, humans have the ability to at least guess the internal states of others under any circumstance. Analyzing the mechanisms by which a human infers information on others and using this mechanism as a clue for creating a framework may be an effective approach. 

Advancing research related to inferring the internal state of a conversational partner not only contributes to the presentation of information to humans at an appropriate granularity, but also to appropriately interpreting information provided by humans and improving the degrees of freedom or efficiency of human-in-the-loop learning. 
True human-agent interactions cannot be established without each inferring the internal state of the other. This should be achieved not by learning each individual's model through a massive set of training data but by developing an effective framework to enable an autonomous agent to use their own experience or the results of estimating some individuals for efficiently estimating the internal state of each person.

\subsection{Consideration of the social aspects of explanation}
As mentioned above, explanations not only play the role of conveying information accurately, but also of changing the beliefs, impressions, and emotions of the person giving the explanation. 
Explanations with the same meaning can result in the loss of a trust relationship between a robot and a human or create a sense of distrust or dislike in the user toward the robot depending on the manner of explanation. 
Therefore, it may be effective when the robot selects an action that the human dislikes for the robot to explain reasons leading up to that behavior or present simulated emotional expressions or words that show a sense of regret. 

Existing research studies have invested significant efforts to minimize the burden of receiving information from a conversation partner by verbalizing and summarizing the explanation; however, there is a lack of discussion of explanation methods. Indeed, it is clear that the realization of appropriate and efficient generation of explanations remains an important research issue; however, clarifying how an explanation affects the relationship between agents and humans and meets the requirements necessary for beneficial co-existence in society are important issues for future research.

\subsection{Application to real-world problems}
Many methods have been proposed that allow autonomous agents and robots that interact with humans to acquire explanatory abilities; however, these methods are presently at the stage of building frameworks limited to a specific domain in a simulation. Application to real-world problems has already begun in the field of XAI, but the area of XAR has currently reached a phase where applications to real-world problems need to be considered.

For example, it is necessary for a robot with a physical body to acquire a world model from partial observation information and predict changes in the external world and its own state with a certain degree of accuracy within that world model. In addition, it is necessary to extract important information for real-world problems that do not generate large differences in action value or to generate explanations using language expressing abstract concepts that cannot be described by an MDP. Developing discussions in a conceptualized field is important to the realization of future interactive agents; however, efforts also need to be invested to introduce the currently proposed framework to real-world problems.

\subsection{Application to interactions}
If we can build an explanation framework that can be applied to real-world problems, it will be thus possible to generate explanations during human–robot interactions. However, it is necessary to determine the timing and content of information presentation autonomously to present an effective explanation during real interactions.

For example, there is no need to explain, even with insufficient comprehension towards the action decisions taken by a robot, if the explanation is not of interest to the user. Alternatively, it is probably better to provide an explanation later if a user is distracted by a separate important activity. Further, there may be scenarios such as XAIP where actions that do not require explanations need to be planned, or XRL where the transparency of the overall policy needs to be ensured. Thus, the timing and content of the explanation to be presented needs to be subdivided further in the context of providing a given explanation; this method of differentiation is not obvious.
	
A useful solution is to define the behavior of presenting an explanation in the action space of the robot. An efficient explanation presentation can be achieved by defining the explanation as an action in line with other actions and by using a conventional learning algorithm without modifications.
To this end, a distinction must be clarified between the internal state of the other, the internal state of the self, and the state of the external world; an effective world model that integrates these three states must also be developed.

\subsection{Ethical issues with generating explanations that are untrue}
When humans explain something to one another, they may often present a process that differs from the actual thought process employed. In such a case, should autonomous robots be allowed to generate such explanations? The ethical aspects of such explanations may be expected to be an important research issue in the future \cite{aborti2019}.

For example, consider a scenario wherein a robot solves a difficult mathematics problem and the user asks how this was done. 
Although the robot implemented a complex formula to solve the problem; it explains the solution to the user by having the user follow a figure to comprehend the solution. 
In this situation, it seems like there is no problem because humans are not particularly harmed. 
Then, can it be problematic in a case such as, for instance, if a robot breaks a cup and purchases a new one, while providing the explanation, ``because the old cup was dirty''? Further, what if it said, ``because today is your birthday''? 
Most humans would are likely to be furious if they found out that a robot was lying. 

As in conversations between humans, it is not necessarily the case that providing accurate explanations will help deepen mutual understanding and build a trust relationship. However, generating only lies would diminish the trust in the explanation itself, and the explanation presentation would lose meaning. To what extent humans can tolerate lies within functional trust relationships, and whether it would even be ethical for a robot to lie in human society remain as important questions that need to be answered to determine how robots should interact in society in the future. These aspects should be actively discussed.



\section{Conclusions}
In this paper, we have first outlined the various types of explainability studied in the field of machine learning to investigate explainability in the context of autonomous robots; then, we proposed a definition for the term. Next, we conducted a survey of research in line with that definition and organized future research issues. The explainability of autonomous robots proposed in this article is important for the social co-existence of autonomous robots to co-exist with humans; however, 
the issues in this field cannot be said to have been researched comprehensively. 
The essence of explainability is communication, and XAR needs to be developed further by integrating machine learning and interaction research. The goal in explainability to define the attitude of ``wanting to explain'' and ``wanting to be understood,'' and ultimately, to convey these feelings through XAR.

\section*{Acknowledgements}
This work was supported by CREST (JPMJCR15E3), JST and the New Energy and Industrial Technology Development Organization (NEDO). 
\bibliographystyle{junsrt}
\bibliography{weekly.bib}






\end{document}